\documentclass{article}

\PassOptionsToPackage{numbers, compress}{natbib}

\usepackage[final]{neurips_2025_ml4ps}
\usepackage[utf8]{inputenc} % allow utf-8 input
\usepackage[T1]{fontenc}    % use 8-bit T1 fonts
\usepackage{hyperref}       % hyperlinks
\usepackage{url}            % simple URL typesetting
\usepackage{booktabs}       % professional-quality tables
\usepackage{amsfonts}       % blackboard math symbols
\usepackage{nicefrac}       % compact symbols for 1/2, etc.
\usepackage{microtype}      % microtypography
\usepackage{xcolor}         % colors
\usepackage{graphicx}
\usepackage{amssymb}
\usepackage{multirow}
 \usepackage{amsmath} 

\title{Variational Autoencoder with Normalizing flow for X-ray spectral fitting}

\author{
    Fiona Redmen \\
    Department of Physics \& Astronomy \\
    University of Southampton \\
    Southampton, SO17 1BJ, UK\\
    \texttt{fr1n21@soton.ac.uk} \\
    \And
    Ethan Tregidga \\
    Laboratoire d’astrophysique \\
    EPFL \\
    Observatoire de Sauverny, 1290 Versoix, Switzerland \\
    \texttt{ethan.tregidga@epfl.ch}\\
    \And
    James Steiner \\
    Harvard-Smithsonian Center for Astrophysics \\
    Cambridge, MA 02138, USA \\
    \texttt{james.steiner@cfa.harvard.edu}\\
    \And
    Cecilia Garraffo \\
    Harvard-Smithsonian Center for Astrophysics \\
    Cambridge, MA 02138, USA \\
    \texttt{cgarraffo@cfa.harvard.edu}
}

\begin{document}

\maketitle

\begin{abstract}
Black hole X-ray binaries (BHBs) can be studied with spectral fitting to provide physical constraints on accretion in extreme gravitational environments. Traditional methods of spectral fitting such as Markov Chain Monte Carlo (MCMC) face limitations due to  computational times. We introduce a probabilistic model, utilizing a variational autoencoder with a normalizing flow, trained to adopt a physical latent space. This neural network produces predictions for spectral-model parameters as well as their full probability distributions. Our implementations result in a significant improvement in spectral reconstructions over a previous deterministic model while performing three orders of magnitude faster than traditional methods.

\end{abstract}

\section{Introduction}
\label{sec:introduction}
Black hole X-ray binaries (BHBs) are composed of a close companion star orbiting, and accreting matter, onto a stellar-mass black hole. The accretion flows in these systems consist of a geometrically-thin and optically-thick disk enshrouded by a corona of hot electrons \citep{remillard_x-ray_2006}. These components emit predominantly in the X-rays, and their resulting spectra can be fit to infer physical properties of the system. Such inference provides key insights into the accretion mechanisms at work \citep{done_modelling_2007}, enabling exploration of General Relativity in extreme environments \citep{fabian_broad_2000}. The standard analytical software for X-ray astronomy,  {\scshape Xspec} \citep{Arnaud1996}, uses Markov Chain Monte Carlo (MCMC) to obtain full posterior distributions on physical parameters. This method of spectral fitting faces limitations due to unwieldy computational times when dealing with a large number of spectra, especially when employing complex models. 

This work advances a new approach, expediting the traditional MCMC method. We use a library of observational data collected for several dozen BHBs with the Neutron-star Interior Composition Explorer (NICER) mission on the International Space Station \citep{gendreau_neutron_2012}, and additionally employ synthetic NICER spectra. Both cover the $0.3-10$ keV energy range with $240$ spectral bins, and from such data we derive full posterior distributions of five physical parameters, along with spectral reconstructions. 

Recent machine learning (ML) work for X-ray astronomy includes representation learning \citep{vagoExtractingLatentRepresentations2025}, with applications in anomaly detection, unsupervised classification, and similarity searches \citep{dillmannRepresentationLearningTimeDomain2025}. Previous ML work in spectral fitting has successfully applied autoencoders (AEs) to this problem, which achieved parameter-value inference $2\,700$ times faster than traditional techniques \citep{tregidga_rapid_2024}. However, that study did not provide uncertainties on the parameter predictions, which is key for scientific interpretability. Moreover, the method required additional iterations of computationally-expensive spectral fitting via {\scshape Xspec} for parameter predictions to remain comparable in accuracy. The works in \citet{barret_simulation-based_2024} and \citet{dupourque_simulation-based_2025} discuss the application of simulation-based inference with automatic posterior transformation \citep{greenbergAutomaticPosteriorTransformation2019} to this problem and accurately obtain full posteriors.

We present an implementation of the network described in \citet{tregidga_rapid_2024} that outputs full posterior distributions. Our posterior predictions reproduce spectra with higher fidelity than before, with similar accuracy to traditional MCMC methods, removing the requirement of additional iterations of {\scshape Xspec} fitting. The architecture takes X-ray spectra as input and outputs the corresponding physical parameter distributions. In addition to maximizing posteriors as in \citet{barret_simulation-based_2024}, we include a decoder, a reconstruction loss, and a latent loss to further modulate training. We demonstrate the robustness of our method when applied to real spectra. The code, which utilizes the repository in \citet{tregidga_rapid_2024}, is available at \url{https://github.com/fi-redmen/fspnet-var.git}.

\section{Method}
\subsection{Datasets}
\label{sec:data}
We aggregate the full library of BHB spectra collected by NICER from July 2017 through August 2022, yielding a total of $10\,800$ spectra, from 25 BHBs. Data is preprocessed as in \citet{tregidga_rapid_2024}. Ground-truth benchmarks, which we call our targets, are obtained as a reference via rigorous fitting of spectra, performed using {\scshape Xspec}. Synthetic spectra are created using {\scshape Xspec}, simulating the physical model with the detector response accounted for, but excluding noise. A total of $100\,000$ spectra are generated, spanning a broad distribution of viable physical parameters. We split the data to train on 80\% of the spectra and use the remaining 20\% as a validation set.

\subsection{Physical Model}
In this exploratory work, we utilize a highly-simplified spectral model which incorporates three spectral features. Thermal emission from the inner thin disk is described by a multicolor black-body continuum \citep{zimmerman_multi-temperature_2005}. The Compton up-scattering of thermal-disk's photons by hot electrons in the corona redistributes a portion of those photons to higher energies, manifesting as a power law \citep{steiner_self-consistent_2017}. Photoelectric absorption by gas along our line of sight to the source cuts off the spectrum at low energies ($\lesssim 1$\;keV) \citep{wilms_absorption_2000}. Through this model, we can determine the following physical properties: temperature of the thin disk $kT_{\mathrm{disk}}$, the amplitude of the thin-disk emission $N$, photon index for the Comptonized emission $\Gamma$, the fraction of photons that are Compton up-scattered $f_{\mathrm{sc}}$, and the line-of-sight column density of neutral gas $N_{\mathrm{H}}$.

\subsection{Architecture}
\label{sec:architecture}

\begin{figure}
  \centering
  \includegraphics[width=\linewidth]{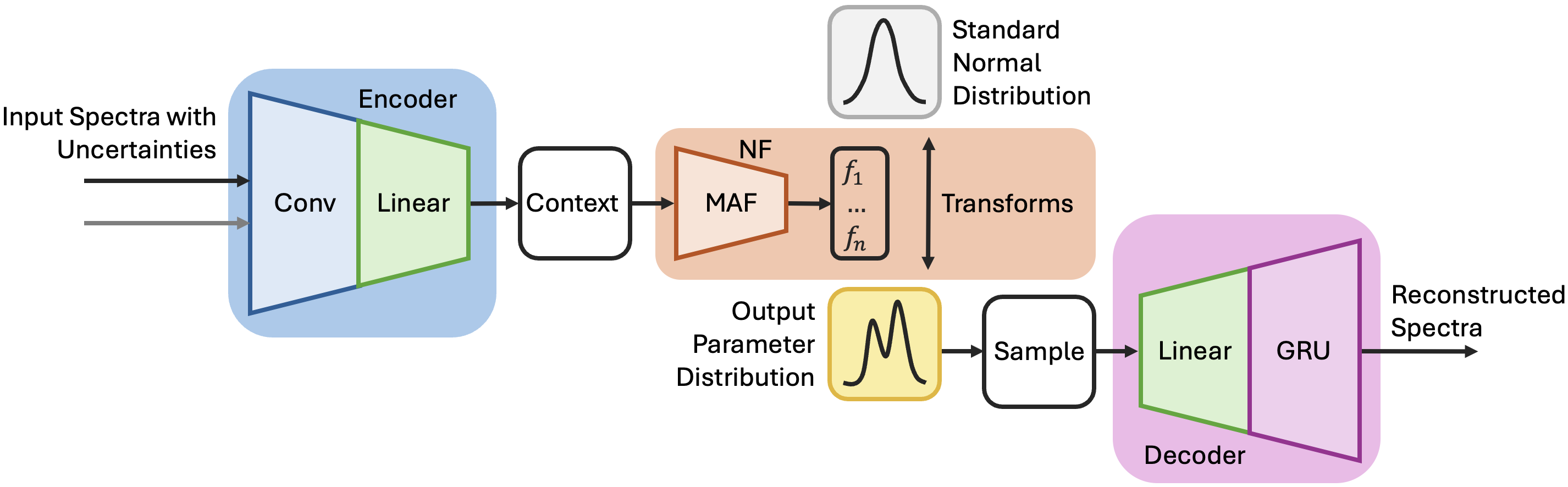}
  \caption{Illustration of AE with NF architecture.}
  \label{fig:network_diagram}
\end{figure}

The network consists of three elements: an encoder, a normalizing flow (NF) \citep{rezende_normalizing_2016}, and a decoder. We use the network and multi-stage training scheme as in \citet{tregidga_rapid_2024}, as this has been shown to effectively perform deterministic spectral parameter predictions. The encoder contains a series of convolutional layers which downscale the dimensionality of the spectra \citep{springenberg_convolutional_2015} and learn patterns in spectra across neighboring data points through various scales. A series of linear layers then allows the encoder to map from the spectral space to the parameter space. The output of the encoder is a partially reduced dimension representation of the spectra, referred to as ``context''. The NF is a neural spline flow \citep{durkan_neural_2019}, parameterized by an autoregressive network \citep{papamakarios_masked_2018}, which outputs a series of ten monotonic rational-quadratic spline transforms \citep{durkan_neural_2019}, conditioned on the context. The transforms describe the predicted posteriors as a map from a standard normal distribution to the posterior. Draws are sampled from the posteriors to form our latent space, and serve as input to the decoder. The decoder follows a similar but reverse-ordered structure to the encoder; a series of linear layers map the parameter space to spectral space and a bidirectional gated recurrent unit \citep{schuster_bidirectional_1997} is used to learn sequential relationships from lower to higher energies and vice versa. We use GELU activation functions throughout the network. The architecture is displayed in Figure~\ref{fig:network_diagram}. For each spectrum, a five-dimensional distribution is predicted, with each dimension representing a different parameter. Using a full NF rather than a vanilla VAE produces complex posteriors which more closely resemble the true spectral parameter distributions.
 
We use a Gaussian negative log likelihood \citep{GNLLL} reconstruction loss, as this prioritizes data points with lower relative errors. Since our goal is to obtain physical parameters, rather than an arbitrary decomposition of the spectra, we impose a physical latent space. To do this, we use an additional latent loss term which forces the latent values to match with physical quantities, penalizing differences as the mean square error. To learn the correct distributions, we include a third flow loss term that maximizes the log probability at the values seen in the dataset with corresponding spectra. The flow loss is mathematically described as

\begin{equation}
    \mathcal{L}_{\text{NF}}(\boldsymbol{\phi}) = -\log q_{\boldsymbol{\phi}}(\boldsymbol{\theta}|\boldsymbol{x})
\end{equation}

\noindent where $q_{\boldsymbol{\phi}}$ describes the predicted probability distribution of parameters $\boldsymbol{\theta}$ given spectra $\boldsymbol{x}$ and $\boldsymbol{\phi}$ represents the network parameters.

The novel technique of implementing a NF within an AE leverages the decoder by informing parameter predictions on degeneracies. For example, the two parameters $N$ and $kT$ both increase with overall spectral flux. Joint over-predictions or under-predictions in $N$ and $kT$ would return inaccurate spectra, penalized strongly through the reconstruction loss. 

We use the AdamW optimizer\citep{loshchilov2019decoupledweightdecayregularization} with an initial learning rate of $1\times10^{-3}$ which is reduced to $1\times10^{-6}$ through a reduce on plateau scheduler. The training is split into three stages. First, the decoder is trained on synthetic data, to learn the physically informed mapping between parameters and spectra. The weights of the decoder are then frozen and the network is trained, end to end, on synthetic data. Finally, the network is trained on real data, with the decoder still frozen, and an initial learning rate of $1\times10^{-4}$. Transfer learning from synthetic to real data allows us to maximize our training samples, while adapting to real data \citep{tan_survey_2018}. Training is halted when a plateau is detected across the last $30$ epochs and capped at $400$ epochs. The three stages are trained for $400$, $121$ and $216$ epochs, respectively. The first stage utilizes only the reconstruction loss, while the last two stages use all three terms with equal weightings. Experimentation shows that reducing the reconstruction loss weight by $1\times10^{-3}$ or removing the latent and reconstruction losses very slightly worsens performance. Longer training of the last stage leads to overfitting past roughly $250$ epochs.

\section{Evaluation}
\label{sec:evaluation}
\begin{figure}
    \centering
    \includegraphics[width=0.92\linewidth]{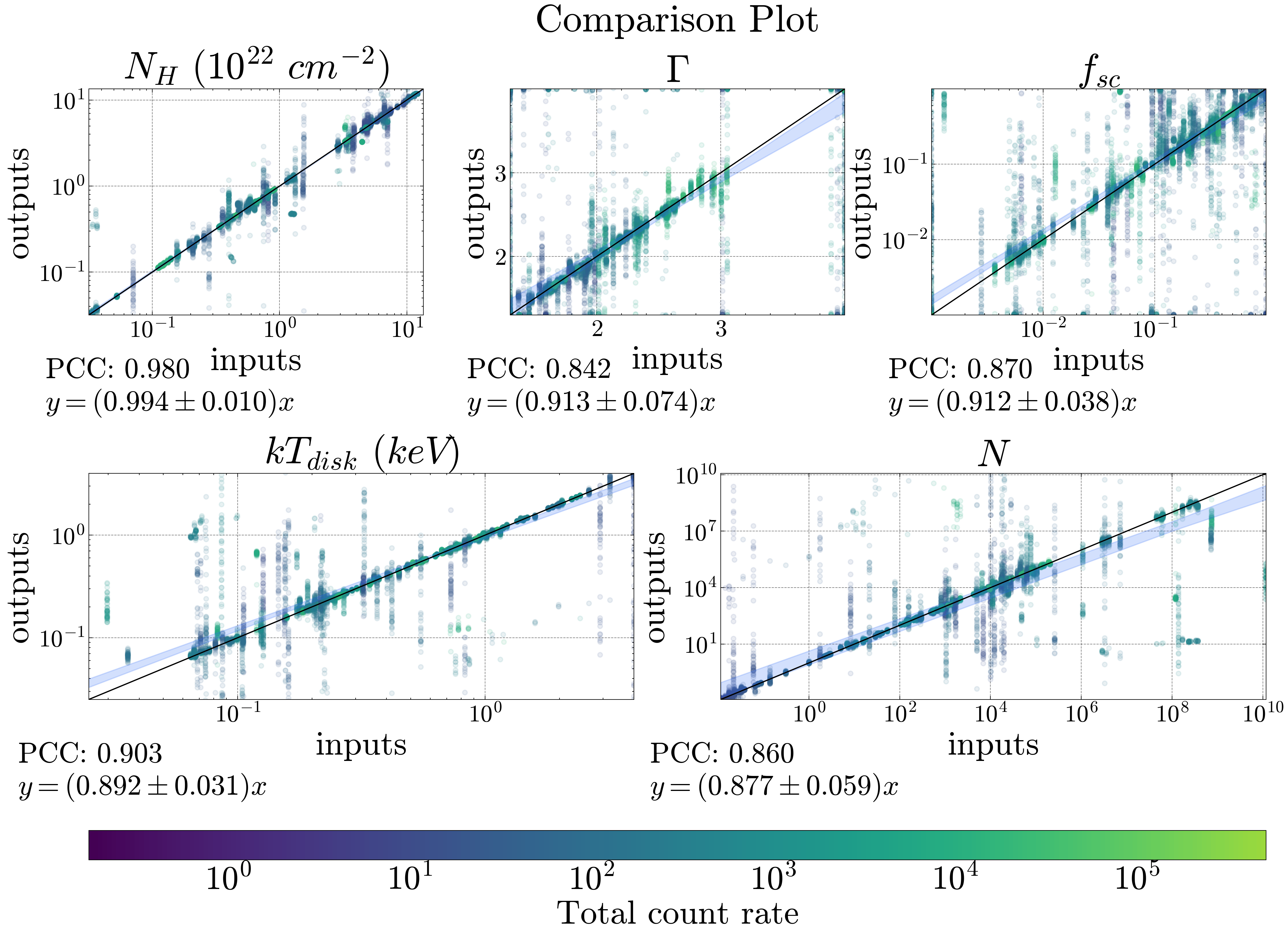}
    \caption{$50$ samples from output posteriors plotted against corresponding input values for a total of $250$ spectra, and colored by total spectral count rate. PCC and best linear fits are included.}
    \label{fig:parameter_comparisons}
\end{figure}

Figure~\ref{fig:parameter_comparisons} shows predicted against target values for a subset of $250$ real spectra. We plot $50$ values sampled from corresponding posteriors to capture the distributions. The data is colored by the total count rate. The Pearson Correlation Coefficient (PCC) between the predicted and target values for each parameter is printed. Best linear regressions are performed over each of the $50$ sets of samples and the mean and standard deviation are taken. We overlay that linear fit and an an idealized line (slope of $1$). There is a strong correlation across all parameters, with a minimum PCC of $0.842$ shown for $\Gamma$. Notably, the spectrum is only weakly affected by $\Gamma$ when either $f_{\rm sc}$ is low or when $kT$ is very large owing to NICER's bandpass. The worst linear fit (slope of $0.877\pm0.059$) is seen for $N$.

The reduced ``PGStat'' \citep{reducedPG} is a fit statistic for Poisson-distributed data with Gaussian-distributed background. An ideal fit would correspond to a reduced PGStat of $1$ and increases with worse fits. We compute the reduced PGStats of reconstructions produced by inputting our parameters to the physical model in \textsc{Xspec}. We take the median value across the set of $2\,160$ spectra in our validation data, for robustness against outliers. The first reduced PGStat value, used as a reference point, is calculated using our target parameters. The second and third values are computed with parameters obtained through \textsc{Xspec} fitting and the NF respectively. Ensuring these values match each other, we allow for a direct comparison in computation time for each method to achieve similar accuracy. \textsc{Xspec} uses $130$ iterations of fitting to fulfill this requirement, obtaining one effective posterior sample. 
The final value, computed using the same network but without the decoder$-$trained on synthetic and real data for 121 and 216 epochs respectively$-$demonstrates the impact of including the decoder. 
To obtain multiple samples in \textsc{Xspec}, we run an MCMC chain.\footnote{
We estimate the computation time for $1000$ effective samples across the $2\,160$ spectra in our validation set by running a representative MCMC chain for one spectrum. We approximate an autocorrelation length, $\tau$, and use  $t_{\rm total} = \frac{(N_{\rm samples} \tau + b)t}{L +b} N_{\rm val}$, where $L$ is the overall chain length, $b$ is the burn in, $t$ is the representative chain run time, $N_{\rm samples}$ is the number of effective samples we aim to obtain, and $N_{\rm val}$ is our validation set size.\label{footnote: MCMC}}
The NF samples are obtained in one forward pass of the network, irrespective of sample size. Computation times were measured over predictions run on an Apple M2 chip.

We show reduced PGStats, along with computational times for parameter predictions over this set of spectra in Table~\ref{table:pg}. Due to the oversimplifications in our model, we do not expect to see a reduced PGStat of $1$ from our pre-calculated targets. The reduced PGStat values resulting from the NF and $130$ fit iterations with \textsc{Xspec} are both comparable to  the reference value. Therefore, the parameters determined by our NF are able to reconstruct spectra through the physical model with similar accuracy to traditional methods, despite the deviations shown from target values in Figure~\ref{fig:parameter_comparisons}. Computing one sampled parameter prediction is $\sim 640$ times faster than \textsc{Xspec} fitting. Computing $1000$ samples for full posterior evaluation is $\sim 2000$ times faster than using MCMC chains. Training the encoder and NF by itself results in worse \textsc{Xspec} reconstructions, indicating that the decoder does inform the physical parameter predictions. The reduced PGStat resulting from the \citet{tregidga_rapid_2024} model is $62.7$ with a baseline value of $4.44$, a factor of $\sim14$ worse than both traditional methods and the model we present.

\begin{table}
  \caption{Reduced PGStats and computation times}
  \label{table:pg}
  \centering
  \begin{tabular}{lll}
    \toprule
    Scenario     & Computation time (s) & Reduced PGStat \\
    \midrule
    Pre-calculated targets              & \multirow{2}{4em}{N/A}                            &\multirow{2}{4em}{$3.801$} \\[4pt] %\\[4pt] %
    (for reference)                     &                                                   &\\[4pt] %\\[4pt] %
    \multirow{2}{4em}{\textsc{Xspec}}   & 1 sample $1279\pm7$                               &\multirow{2}{4em}{$3.804$} \\[4pt] %\\[4pt] %
                                        & 1000 samples: $\sim100\,000$~\ref{footnote: MCMC} &\\[4pt] \\[4pt] %
    \multirow{2}{4em}{NF in AE}         & 1 sample: $2.11\pm0.05$                           &\multirow{2}{4em}{$3.796$}  \\[4pt] %\\[4pt] %
                                        & 1000 samples: $51.8\pm0.9$                        & \\[4pt] %\\[4pt] %
    NF                                  & Same as NF in AE                                  & $6.034$ \\[4pt] %\\[4pt] %
    \bottomrule
  \end{tabular}
\end{table}

\begin{figure}
    \centering
    \includegraphics[width=0.35\columnwidth]{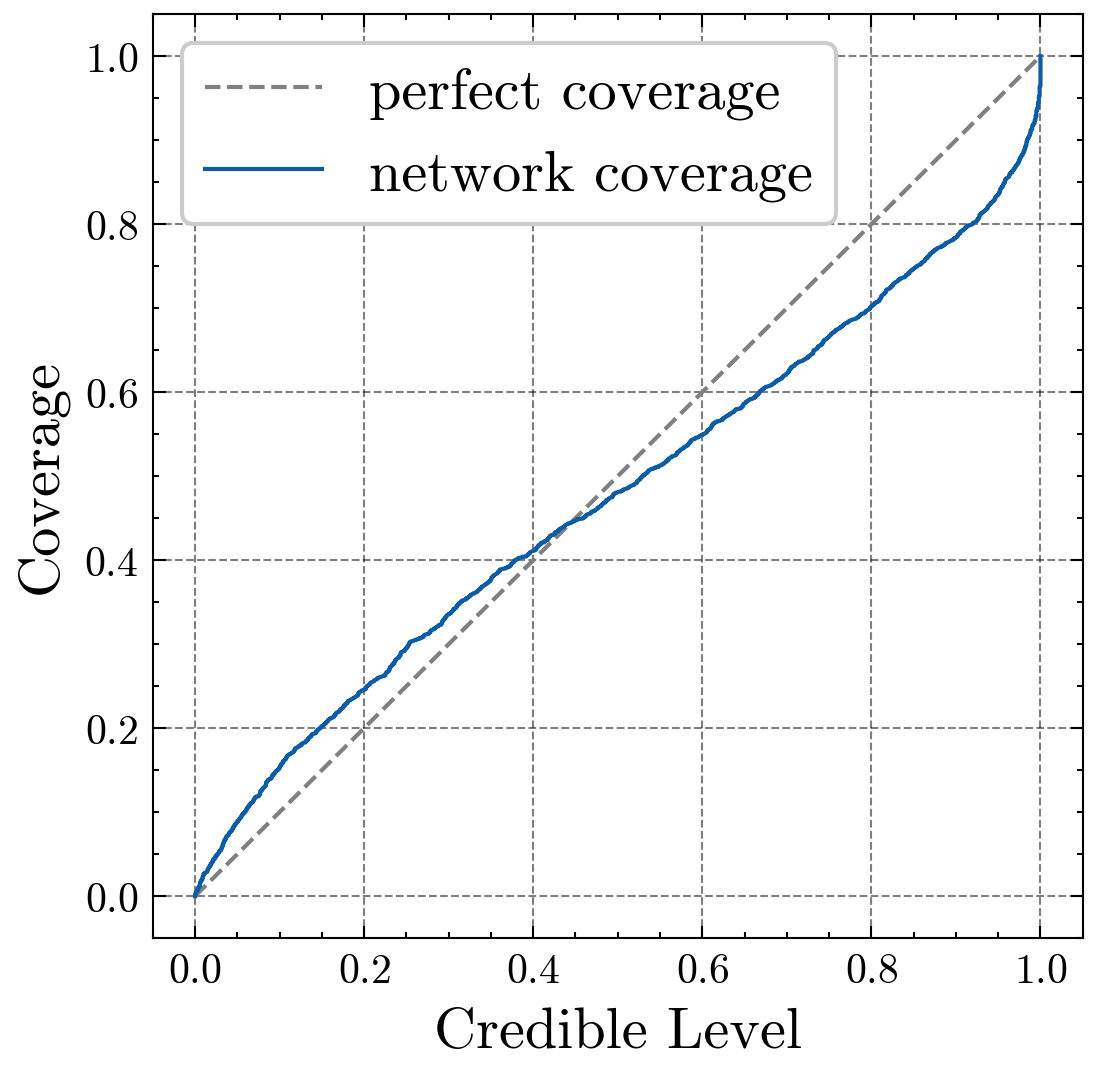}
\caption{Plot of coverage versus credible level for distributions predicted from real spectra (blue, solid). The ideal coverage is also plotted (grey, dashed).}
\label{fig:real_coverage}
\end{figure}

Fig.~\ref{fig:real_coverage} is a coverage plot for distributions predicted from real data. At low credible levels, the network is slightly under-confident, while at high credible levels it is slightly over confident. 

\section{Conclusion}
\label{sec:conclusion}
This work describes the implementation of a NF in an AE network to infer full posterior predictions of physical parameters from BHB spectra. We demonstrate the robustness of this method by performing evaluations on real data predictions. The NF implementation obtains accurate parameter predictions, with most inaccuracies and imprecisions attributable to low count rates in spectra. While remaining comparable in accuracy, posterior predictions are $\sim 640$ times faster than $130$ iterations of \textsc{Xspec} fitting. This computational speed-up is extended dramatically to a factor of $\sim2000$ when comparing to MCMC posterior estimation in \textsc{Xspec}.

This method can already be applied to perform population studies across the NICER archive, however, further training optimization may improve performance, including more thorough investigations into loss function weightings. Currently, our target parameters are determined using an oversimplified model in \textsc{Xspec}, limiting the accuracy of network predictions versus true physical values. Future work could apply the network on more complex, physically accurate models, which also contain more information on physics within the systems.

\begin{ack}
% Use unnumbered first level headings for the acknowledgments. All acknowledgments

Thanks to Diego Altamirano, Joshua Wing, Rafael Martínez-Galarza and the anonymous referees for their advice and support. Additionally, thanks to the Astrophysics with year Abroad program at the University of Southampton. This project was also conducted with support from the AstroAI initiative at the Center for Astrophysics | Harvard \& Smithsonian. This research has made use of data and/or software provided by the High Energy Astrophysics Science Archive Research Center (HEASARC), which is a service of the Astrophysics Science Division at NASA/GSFC. E.T. and J.F.S. acknowledge support from NASA grant no. 80NSSC21K1886.

% Do {\bf not} include this section in the anonymized submission, only in the final paper. You can use the \texttt{ack} environment provided in the style file to automatically hide this section in the anonymized submission.
\end{ack}

% \section*{References}

% References follow the acknowledgments in the camera-ready paper. Use unnumbered first-level heading for
% the references. Any choice of citation style is acceptable as long as you are
% consistent. It is permissible to reduce the font size to \verb+small+ (9 point)
% when listing the references.
% Note that the Reference section does not count towards the page limit.
% \medskip

\newpage
{
\small

\bibliography{refs}

}

%%%%%%%%%%%%%%%%%%%%%%%%%%%%%%%%%%%%%%%%%%%%%%%%%%%%%%%%%%%%

% \appendix

% \section{Technical Appendices and Supplementary Material}
% Technical appendices with additional results, figures, graphs and proofs may be submitted with the paper submission before the full submission deadline (see above), or as a separate PDF in the ZIP file below before the supplementary material deadline. There is no page limit for the technical appendices.

\end{document}